\documentclass{article}
\usepackage{times}
\usepackage{amsmath,amsfonts,epsfig}
\newcommand{\myvcenter}[1]{\ensuremath{\vcenter{\hbox{#1}}}}
\newcommand{\cind}{\perp\hspace*{-1.35ex}\perp}

\title{Mixed Cumulative Distribution Networks}

\author{
       Ricardo Silva \\
       {\tt \small ricardo@stats.ucl.ac.uk} \\
       Department of Statistical Science
       University College London  \\ \\
       Charles Blundell \\
       {\tt \small c.blundell@gatsby.ucl.ac.uk} \\
       Gatsby Computational Neuroscience Unit\\ 
       University College London \\ \\
       Yee Whye Teh \\
       {\tt \small ywteh@gatsby.ucl.ac.uk}\\
       Gatsby Computational Neuroscience Unit\\
       University College London}

\begin{document}

\maketitle

\begin{abstract}
  Directed acyclic graphs (DAGs) are a popular framework to express
  multivariate probability distributions. Acyclic directed mixed
  graphs (ADMGs) are generalizations of DAGs that can succinctly
  capture much richer sets of conditional independencies, and are
  especially useful in modeling the effects of latent variables
  implicitly. Unfortunately there are currently no good
  parameterizations of general ADMGs. In this paper, we apply recent
  work on cumulative distribution networks and copulas to propose one
  general construction for ADMG models. We consider a simple parameter estimation
  approach, and report some encouraging experimental results.
\end{abstract}

\section{Contribution}

Graphical models provide a powerful framework for encoding
independence constraints in a multivariate distribution
\cite{pearl:88,lauritzen:96}. Two of the most common families, the
directed acyclic graph (DAG) and the undirected network, have
complementary properties. For instance, DAGs are non-monotonic
independence models, in the sense that conditioning on extra variables
can also destroy independencies (sometimes known as the ``explaining
away'' phenomenon \cite{pearl:88}). Undirected networks allow for
flexible ``symmetric'' parameterizations that do not require a
particular ordering of the variables.

\begin{figure}
\begin{center}
\begin{tabular}{cc}
\myvcenter{\epsfig{file=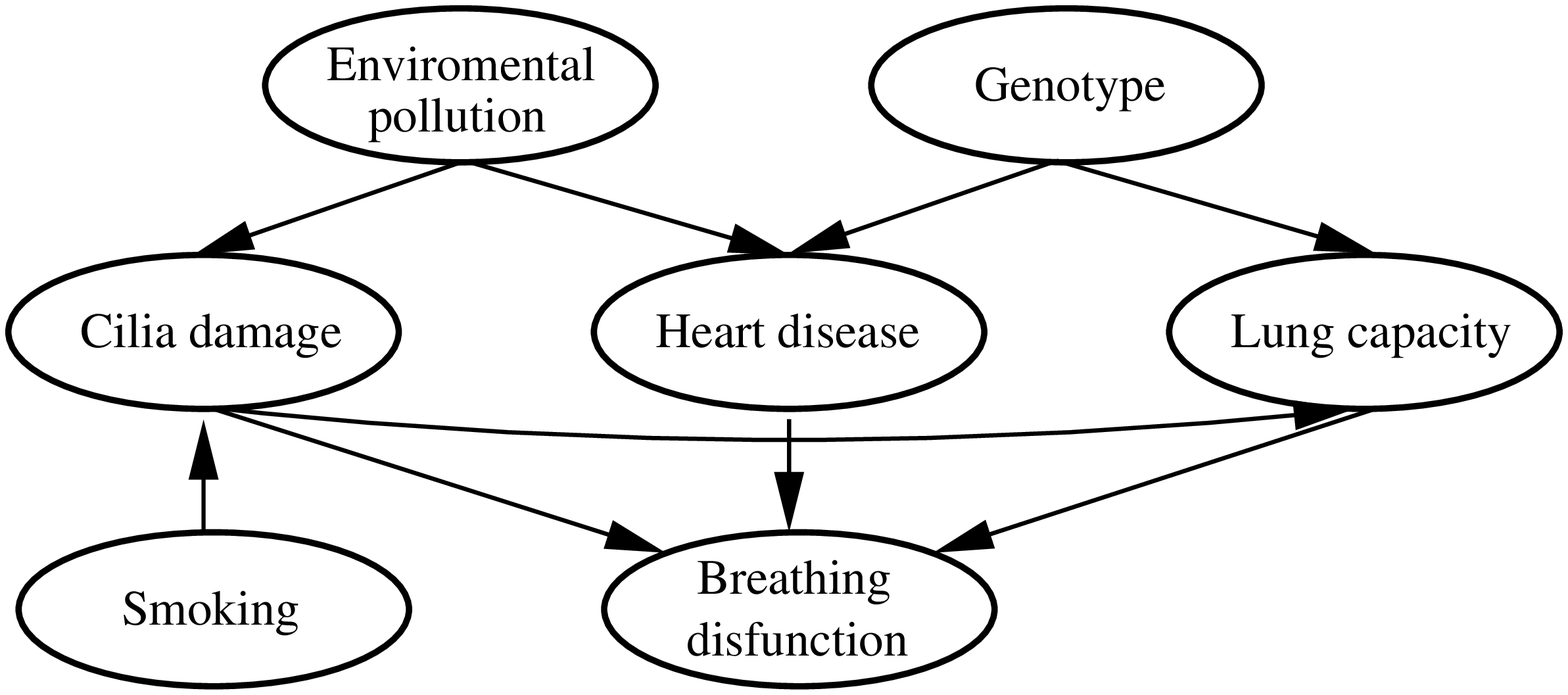,width=6.5cm}} &
\myvcenter{\epsfig{file=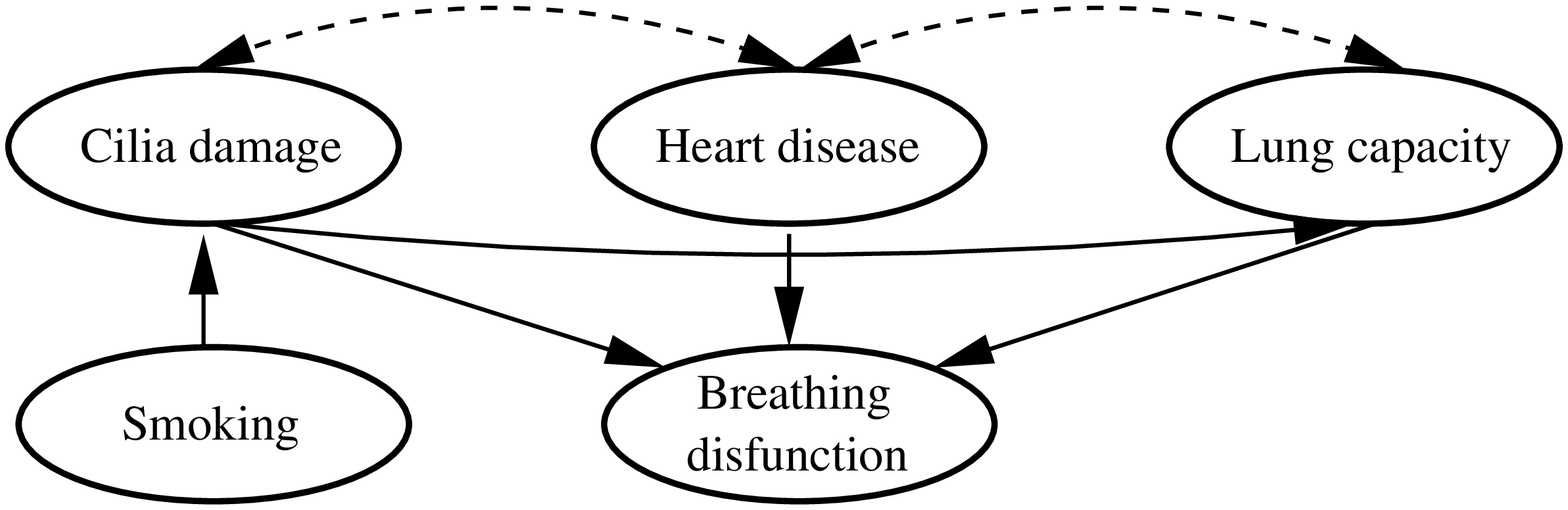,width=6.5cm}} \\
(a) & (b))\\
\end{tabular}
\end{center}
\caption{(a) A DAG representing dependencies over a set of variables
(adapted from \cite{sgs:00}, page 137) in a medical domain.
(b) The ADMG representing conditional independencies corresponding to
(a), but only among the remaining vertices: pollution and genotype factors were
marginalized. In general, bi-directed edges emerge from unspecified variables that
have been marginalized but still have an effect on the remaining variables. The ADMG is acyclic in the sense
that there are no cycles composed of directed edges only. In general, a DAG cannot
represent the remaining set of independence constraints after some
variables in another DAG have been marginalized.}
\label{fig:example}
\end{figure}

More recently, alternative graphical models that allow for both
directed and symmetric relationships have been introduced. The {\it
  acyclic directed mixed graph} (ADMG) has both directed and
bi-directed edges and it is the result of {\it marginalizing} a DAG:
Figure \ref{fig:example} provides an example.
\cite{richardson:02,richardson:03} show that DAGs are not closed under
marginalization, but ADMGs are. Reading off independence constraints
from a ADMG can be done with a procedure essentially identical to
d-separation \cite{pearl:88, richardson:02}.

Theoretical properties and practical applications of ADMGs are further
discussed in detail by e.g.
\cite{bol:89,sgs:00,drton:08,zhang:08,pellet:08,silva:09,huang:10}.
One can also have latent variable ADMG models, where bi-directed edges
represent a subset of latent variables that have been marginalized. In
sparse models, using bi-directed edges in ADMGs frees us from having
to specify exactly which latent variables exist and how they might be
connected. In the context of Bayesian inference, Markov chain Monte
Carlo in ADMGs might have much better mixing properties compared to
models where all latent variables are explicitly included
\cite{silva:09}.

However, it is hard in general to parameterize a likelihood function
that obeys the independence constraints encoded in an ADMG. Gaussian
likelihood functions and their variations (e.g., mixture models and
probit models) have been the only families exploited in most of the
literature \cite{richardson:02,silva:09}. The contribution of this
paper is to provide a flexible construction procedure to design
probability mass functions and density functions that are Markov with
respect to an arbitrary ADMG.  This is done by exploiting recent work
on \emph{cumulative distribution networks} \cite{huang:08} and
\emph{copulas} \cite{nelsen:07,kirshner:07}.  We also provide a
straightforward approach to learning in our ADMGs inspired by the
parameter estimation approaches in the copula literature.  We review
mixed graphs and cumulative distribution networks in Section
\ref{sec:cdns}. The full formalism is given in detail in Section
\ref{sec:mcdn}.  An instantiation of the framework based on copulas
and a parameter estimation procedure is described in Section
\ref{sec:copula}.  Experiments are described in Section
\ref{sec:experiments}, and we conclude with Section
\ref{sec:conclusions}.

\section{Mixed Graphs and Cumulative Distribution Networks}
\label{sec:cdns}

In this section, we provide a summary of the relevant properties of
mixed graph models and cumulative distribution networks, and the
relationship between formalisms.

A {\it bi-directed} graph is a special case of a ADMG without directed
edges.  The absence of an edge $(X_i, X_j)$ implies that $X_i$ and
$X_j$ are {\it marginally independent}. Hence, bi-directed models are
{\it models of marginal independence} \cite{drton:08}. Just like in a
DAG, conditioning on a vertex that is the endpoint of two arrowheads
will make some variables dependent.  For instance, for a bi-directed
graph $X_1 \leftrightarrow X_2 \leftrightarrow X_3$, we have that
$X_1\cind X_3$ but $X_1\not\cind X_3|X_2$.  See
\cite{drton:03,drton:08} for a full discussion.

Current parameterizations of bi-directed graphs suffer from a number
of practical difficulties.  For example, consider binary bi-directed
graphs, where a complete parameterization was introduced by Drton and
Richardson \cite{drton:08}. Let $\mathcal G$ be a bi-directed graph
with vertex set $X_V$. Let $q_A \equiv P(X_A = 0)$, for any vertex set
$X_A$ contained in $X_V$. The joint probability $P(X_A = 0, X_{V
  \backslash A} = 1)$ is given by
\begin{equation}
\label{eq:mobius}
P(X_A = 0, X_{V \backslash A} = 1) = \sum_{B:A \subseteq B} (-1)^{|B\backslash A|}q_B
\end{equation}
The set $\{q_S:S \subset S\}$ is known as the M\"{o}bius
parameterization of $P(X_V)$, since relationship (\ref{eq:mobius}) is
an instance of the M\"{o}bius inversion operation \cite{lauritzen:96}.
The marginal independence of the bi-directed graph implies $P(X_A = 0,
X_B = 0) = P(X_A = 0)P(X_B = 0)$ if no element in $X_A$ is adjacent to
any element in $X_B$ in $\mathcal G$. Therefore, the set of
independent parameters in this parameterization is given by $\{q_A\}$,
for all $X_A$ that forms a connected set in $\mathcal G$. This
parameterization is complete, in the sense that {\it any} binary model
that is Markov with respect to $\mathcal G$ can be represented by the
set $\{q_A\}$. However, this comes at a price: in general, the number
of connected sets can grow exponentially in $|X_V|$ even for a sparse,
tree-structured, graph. Moreover, the set $\{q_A\}$ is not {\it
  variation independent} \cite{lauritzen:96}: the parameter space is
defined by exponentially many constraints. In contrast, different
conditional probability tables in a given Bayesian network can be
parameterized independently \cite{lauritzen:96,pearl:88}.

Cumulative distribution networks (CDNs), introduced by Huang and Frey
\cite{huang:08} as a convenient family of cumulative distribution
functions (CDFs), provide a alternative construction of bi-directed
models by indirectly introducing additional constraints to reduce the
total number of parameters. Let $X_V$ be a set of random variables,
and let $\mathcal G$ be a bi-directed graph\footnote{\cite{huang:08}
  describe the model in terms of factor graphs, but for our purposes a
  bi-directed representation is more appropriate.}  with $\mathcal C$
being a set of cliques in $\mathcal G$.  The CDF over $X_V$ is given
by
\begin{equation}
\label{eq:cdn}
 P(X_V \le x_V) \equiv F(x_V) = \prod_{S \in \mathcal C}F_S(x_S)
\end{equation}
\noindent where each $F_S$ is a parametrized CDF over $X_S$. A
sufficient condition for (\ref{eq:cdn}) to define a valid CDF is that
each $F_S$ is itself a CDF. CDNs satisfy the conditional independence
constraints of bi-directed graphs \cite{huang:08}.  For example,
consider $X_1 \leftrightarrow X_2 \leftrightarrow X_3$, with cliques
$X_{S_1} = \{X_1, X_2\}$ and $X_{S_2} = \{X_2, X_3\}$. The marginal
CDF of $X_1$ and $X_3$ is $P(X_1 \le x_1, X_3 \le x_3)=P(X_1 \le x_1,
X_2 \le \infty, X_3 \le x_3) = F_1(x_1, \infty)F_2(\infty, x_3)$.
Since this factorizes, it follows that $X_1$ and $X_3$ are marginally
independent.


The relationship between the complete parameterization of Drton and
Richardson and the CDN parameterization can be exemplified in the
discrete case. Let each $X_i$ take values in $\{0, 1, 2, ...\}$.
Recall that the relationship between a CDF and a probabiliy mass
function is given by the following inclusion-exclusion formula
\cite{joe:97}:
\begin{equation}
\label{eq:cdf2pmf}
P(x_1, \dots, x_{d}) = \sum_{z_1 = 0}^{1}\sum_{z_2 = 0}^{1}\dots\sum_{z_{d} = 0}^{1}
(-1)^{z_1 + z_2 + \dots z_{d}}F(x_{1 - z_1}, x_{2 - z_2}, \dots,
x_{d - z_{d}}),
\end{equation}
\noindent for $d = |X_V|$. In the binary case, since $q_A = P(X_A = 0) = P(X_A \le
0, X_{V \backslash A} \le 1) = F(x_A = 0, x_{V \backslash A} = 1)$, one
can check that (\ref{eq:cdf2pmf}) and (\ref{eq:mobius}) are the
same expression. The difference between the CDN parameterization \cite{huang:08}
and the complete parameterization \cite{drton:08} is that, on top of enforcing
$q_{A \cup B} = q_Aq_B$ for $X_A$ disconnected from $X_B$, we have the additional
constraints
\begin{equation}
\label{eq:min_constraint}
q_A = \prod_{A_C \in \mathcal C(A)} q_{A_C}
\end{equation}
for each connected set $X_A$, where $\mathcal C(A)$ are the maximal
cliques in the subgraph obtained by keeping only the vertices
$X_A$ and the corresponding edges from $\mathcal G$\footnote{This property was
called {\it min-independence} in \cite{huang:09}.}. 

As a framework for the construction of bi-directed models, CDNs have
three major desirable features. Firstly, the number of parameters
grows with the size of the largest clique, instead of $|X_V|$.
Secondly, parameters in different cliques are variation independent,
since (\ref{eq:cdn}) is well-defined if each individual factor is a
CDF. Thirdly, this is a general framework that allows not only for
binary variables, but continuous, ordinal and unbounded discrete
variables as well.  Finally, in graphs with low tree-widths,
probability densities/masses can be computed efficiently by dynamic
programming \cite{huang:08}.
To summarize, CDNs provide a restricted family of marginal
independence models, but one that has computational, statistical and modeling
advantages. Depending on the application, the extra constraints are
not harmful in practice, as demonstrated by \cite{huang:10}.

\section{Mixed Cumulative Distribution Models}
\label{sec:mcdn}

In what follows, we will extend the CDN family to general acyclic
directed mixed graphs: the {\it mixed} cumulative distribution network
(MCDN) model.  In Section \ref{sec:highlevel}, we describe a
higher-level factorization of the {\it probability} (mass or density)
{\it function} $P(X_V)$ involving subgraphs of $\mathcal G$. In
Section \ref{sec:barren}, we describe cumulative distribution
functions that can be used to parameterize each factor defined in
Section \ref{sec:highlevel}, in the special case where no directed
edges exist between members of a same subgraph.  Finally, in Section
\ref{sec:general}, we describe the general case.

\begin{figure}
\begin{center}
\begin{tabular}{ccc}
\myvcenter{\epsfig{file=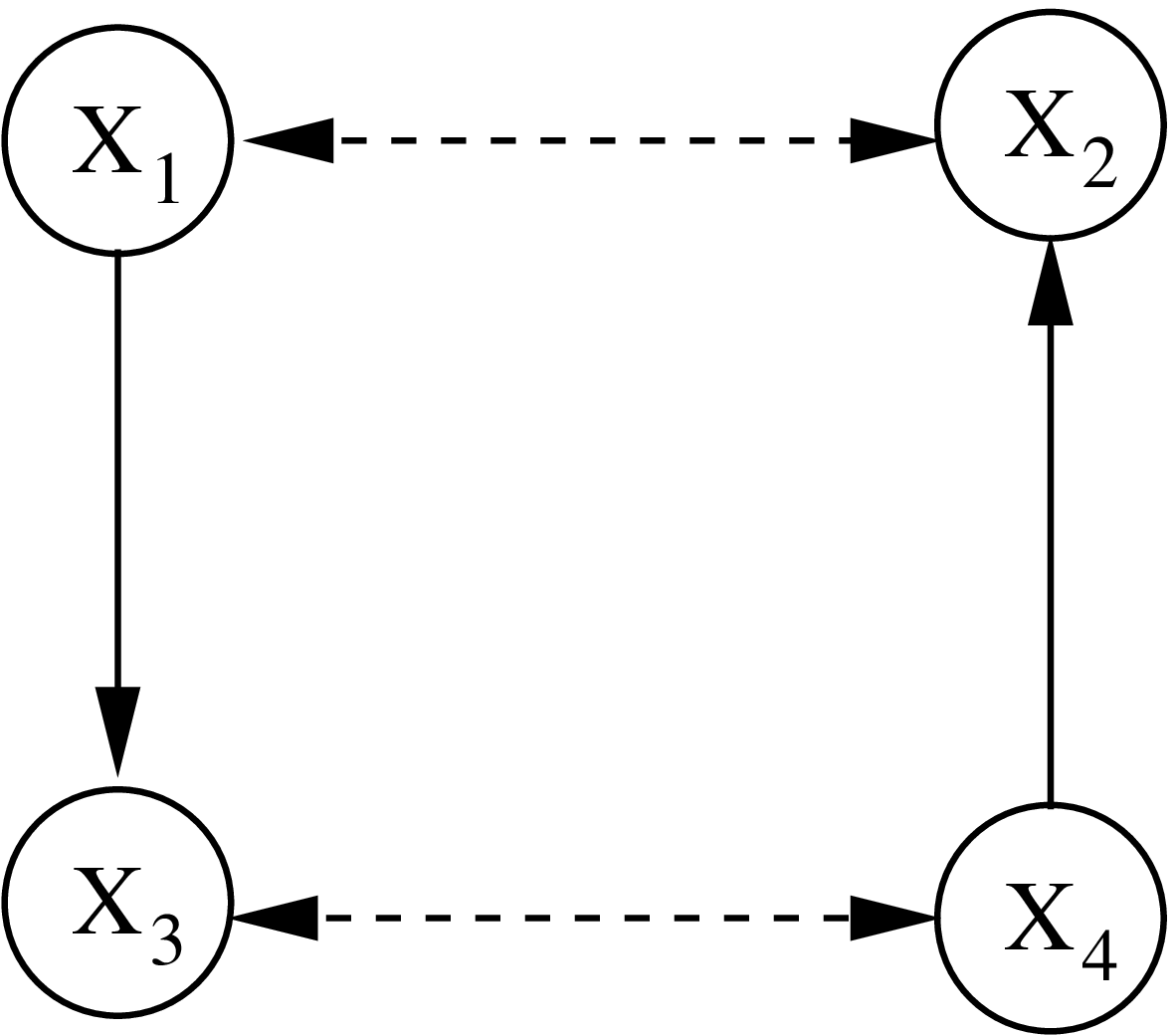,width=3.3cm}} &
\myvcenter{\epsfig{file=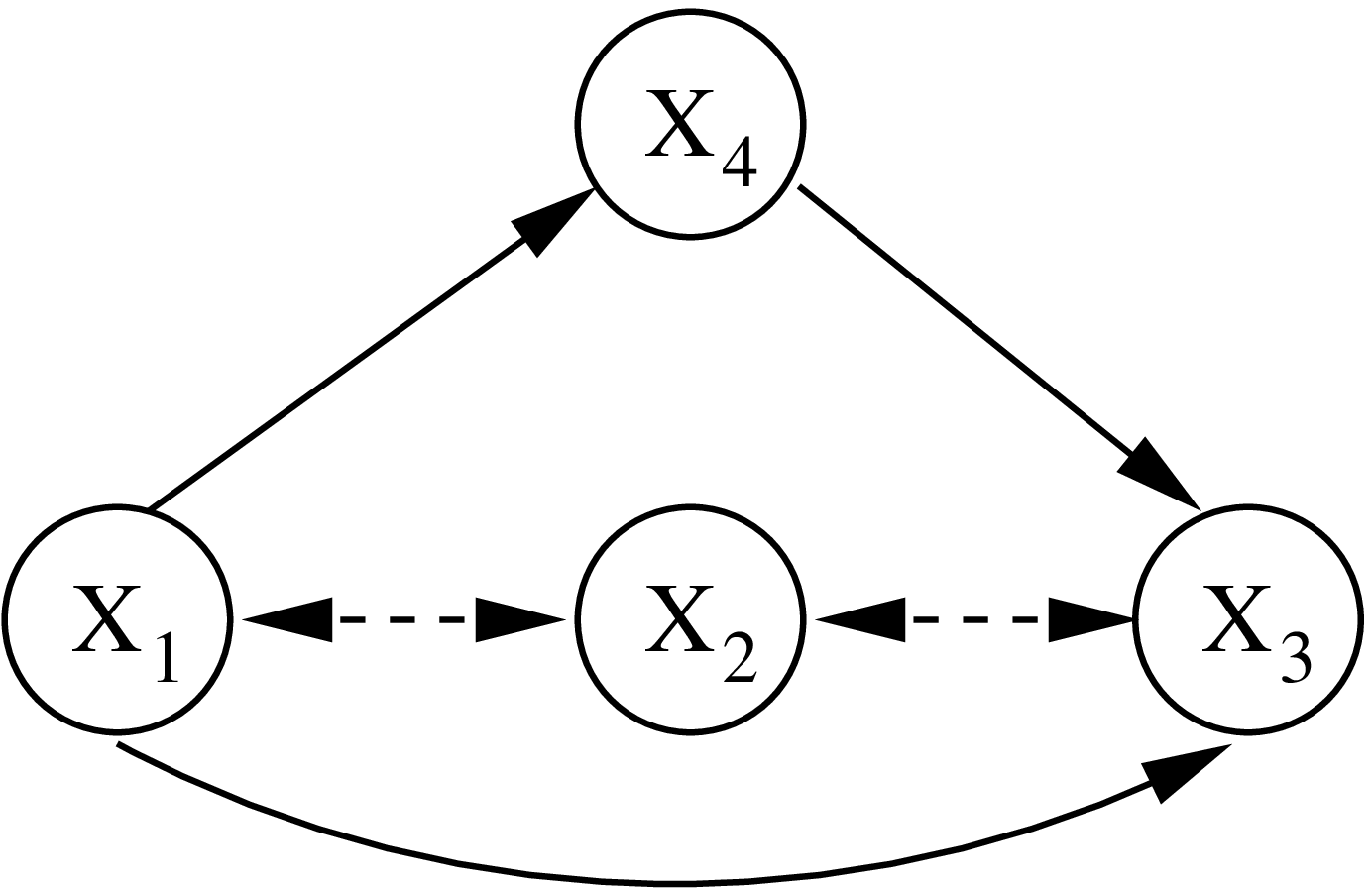,width=4.3cm}} &
\myvcenter{\epsfig{file=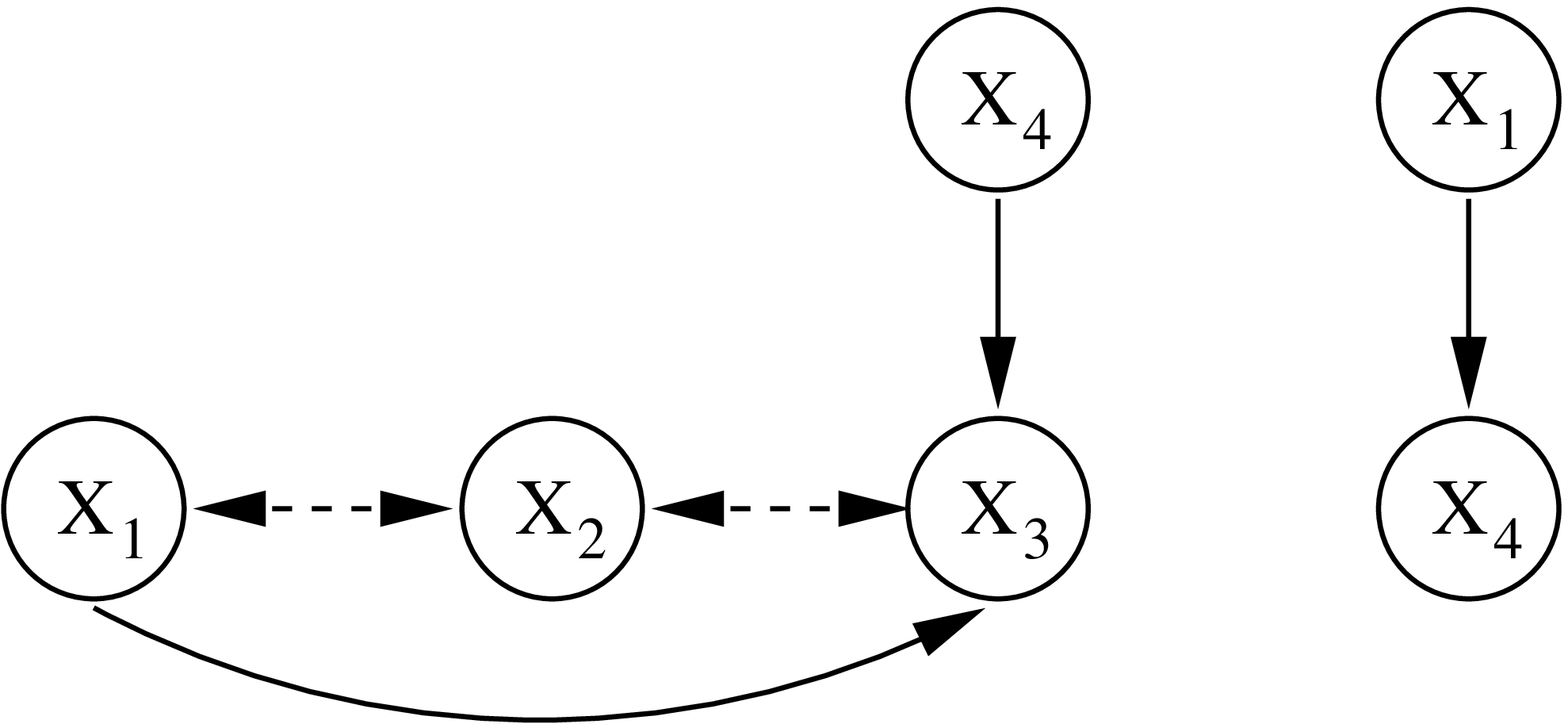,width=5.0cm}} \\
(a) & (b) & (c)\\
\end{tabular}
\end{center}
\caption{(a) The ADMG has two districts, $X_{D_1} = \{X_1, X_2\}$ with singleton parent
$X_4$, and $X_{D_2} = \{X_3, X_4\}$ with parent $X_1$.
(b) A more complicated example with two districts. Notice that the district given by $X_{D_1} = 
\{X_1, X_2, X_3\}$ has as external parent $X_4$, but internally some members of the
district might be parents of other members. The other district is a singleton,
$X_{D_2} = \{X_4\}$. (c) The two corresponding subgraphs $\mathcal G_1$
and $\mathcal G_2$ are shown here.}
\label{fig:district}
\end{figure}

Some important notation and definitions: there are two kinds of edges
in an ADMG; either $X_k\rightarrow X_j$ or $X_k\leftrightarrow X_j$.
In the former case (but not the latter) we call $X_k$ a parent of
$X_j$.  We use $pa_{\mathcal G}(X_A)$ to represent the parents of a
set of vertices $X_A$ in graph $\mathcal G$.  For a given $\mathcal
G$, $(\mathcal G)_A$ represents the subgraph obtained by removing from
$\mathcal G$ any vertex {\it not} in set $A$ and the respective edges;
$(\mathcal G)_{\leftrightarrow}$ is the subgraph obtained by removing
all directed edges.  We say that a set of nodes $A$ in $\mathcal G$ is
an {\it ancestral set} if it is closed under the ancestral
relationship: if $X_v \in A$, then all ancestors of $X_v$ in $\mathcal
G$ are also in $A$. Finally, define the districts of a graph $\mathcal
G$ as the connected components of $(\mathcal G)_\leftrightarrow$.
Hence each district is a set of vertices, $X_D$, such that if $X_i$
and $X_j$ are in $X_D$ then there is a path connecting $X_i$ and $X_j$
composed entirely of bi-directed edges.  Note that trivial districts
are permitted, where $X_D = \{X_i\}$.  Associated with each district
$X_{D_i}$ is a subgraph ${\mathcal G}_i$ consisting of nodes
$X_{D_i}\cup pa_{\mathcal G}(X_{D_i})$.  The edges of $\mathcal G_i$
are all of the edges of $(\mathcal G)_{X_{D_i}\cup pa_{\mathcal
    G}(X_{D_i})}$ excluding all edges among $pa_{\mathcal G}(X_{D_i})
\backslash X_{D_i}$.  Two examples are shown in Figure
\ref{fig:district}.

\subsection{District factorization}
\label{sec:highlevel}

Given any ADMG $\mathcal G$ with vertex set $X_V$, we parameterize its  probability mass/density function  as:
\begin{equation}
\label{eq:factorization}
P(X_V) = \prod_{i = 1}^K P_i(X_{D_i}\ |\ pa_{\mathcal G}(X_{D_i}) \backslash X_{D_i})
\end{equation}
\noindent where $\{X_{D_1}, X_{D_2}, \dots, X_{D_K}\}$ is the set of
districts of $\mathcal G$. That is, each factor is a probability
(mass/density) function for $X_{D_i}$ given its set of parents in
$\mathcal G$ (that are not already in $X_{D_i}$). We require that
\begin{itemize}
\item Each $P_i(X_{D_i}\ |\ pa_{\mathcal G}(X_{D_i}) \backslash X_{D_i})$ is Markov with respect to
$\mathcal G_i$,
\end{itemize}
\noindent where a probability function $P(\cdot)$ is {\it Markov with
respect} to a ADMG $\mathcal G$ if any conditional independence
constraint encoded in $\mathcal G$ is exhibited 
in $P(\cdot)$. 

The relevance of this factorization is summarized by the following result.\\

\noindent {\bf Proposition 1.} {\it A probability function $P(X_V)$ is Markov
with respect to $\mathcal G$ if it can be factorized according to
(\ref{eq:factorization}) and each $P_i(X_{D_i}\ |\
pa_{\mathcal G}(X_{D_i}) \backslash X_{D_i})$ is Markov with respect
to the respective $\mathcal G_i$.} \\

Proofs of all results are in Appendix A. 

Note that (\ref{eq:factorization}) is seemingly cyclical: for
instance, Figure \ref{fig:district}(a) implies the factorization
$P_1(X_1, X_2\ |\ X_4)P_2(X_3, X_4\ |\ X_1)$. This suggests that there
are additional constraints tying parameters across different factors.
However, there are no such constraints, as guaranteed through the
following result: \\

\noindent {\bf Proposition 2.} {\it Given an ADMG $\mathcal G$ with respective
subgraphs $\{\mathcal G_i\}$ and districts $\{X_{D_i}\}$, any collection of probability functions
$P_i(X_{D_i}\ |\ pa_{\mathcal G}(X_{D_i}) \backslash X_{D_i})$, Markov
with respect to the respective $\mathcal G_i$, implies that (\ref{eq:factorization})
is a valid probability function (a non-negative function that integrates to 1).}\\

The implication is that one can independently parameterize each
individual $P_i(\cdot\ |\ \cdot)$ to obtain a valid $P(X_V)$ Markov
with respect to any given ADMG $\mathcal G$. In the next sections, we
show how to parameterize each $P_i(\cdot\ | \cdot)$ by factorizing its
corresponding cumulative distribution function.

\subsection{Models with barren districts}
\label{sec:barren}

Consider first the case where district $X_{D_i}$ is {\it barren}, that
is, no $X_v \in X_{D_i}$ has a parent also in $X_{D_i}$
\cite{richardson:09}.  For a given $\mathcal G_i$ with respective
district $X_{D_i}$, consider the following function:
\begin{equation}
\label{eq:barren_fact}
\displaystyle
F_{i}(x_{D_i}\ |\ pa_{\mathcal G}(X_{D_i})) \equiv 
           \left[\prod_{X_S \in \mathcal C_i} F_S(x_S\ |\ pa_{\mathcal G}(X_{D_i}))\right]
           \left[\prod_{X_v \in X_{D_i}}F_v(x_v\ |\ pa_{\mathcal G}(X_v))\right]
\end{equation}
\noindent where $\mathcal C_i$ is the set of cliques in $(\mathcal
G_i)_{\leftrightarrow}$. Each term on the right hand side is a
conditional cumulative distribution function: for sets of random
variables $Y$ and $Z$, $F(y\ |\ z) \equiv P(Y \le y\ |\ Z = z)$. \\

\noindent {\bf Proposition 3.} {\it $F_{i}(x_{D_i})$ is a CDF for any choice of
$\{\{F_S(x_S)\}, \{F_v(x_v\ |\ pa_{\mathcal G}(X_v))\}\}$.
If, according to each $F_S(x_S)$, $X_s \in X_S$ is marginally
independent of any element in
$pa_{\mathcal G}(X_{D_i})\backslash pa_{\mathcal G}(X_s)$, the
corresponding conditional probability function 
$F_{i}(x_{D_i}\ |\ pa_{\mathcal G}(X_{D_i}))$ is Markov with respect to
$\mathcal G_i$.}\\

%

Notice that the structure of type IV chain graphs \cite{drton:09} is a
special case of ADMGs with barren districts. The parameterization of
\cite{drton:09} is complete for such graph models, but requires
exponentially many parameters even in sparse models.

To obtain the probability function (\ref{eq:factorization}), we
calculate each $P_i(X_{D_i}\ |\ pa_{\mathcal G}(X_{D_i})\backslash
X_{D_i} )$ by differentiating the corresponding (\ref{eq:barren_fact})
with respect to $X_{D_i}$. Although this operation, in the discrete
case, is in the worst-case exponential in $|X_{D_i}|$, it can be
performed efficiently for graphs where $(\mathcal
G)_{\leftrightarrow}$ has low tree-width \cite{huang:08}.

\subsection{The general case: reduction to barren case}
\label{sec:general}

We reduce graphs with general districts to graphs with only barren
districts by introducing artificial vertices. Create a graph $\mathcal
G^\star$ with the same vertex set as $\mathcal G$ and the same
bi-directed edges.  For each vertex $X_v$ in $\mathcal G$, perform the
following operation:
\begin{itemize}
\item add an artificial vertex $X_v^\star$ to $\mathcal G^\star$;
\item add the edge $X_v \rightarrow X_v^\star$ to $\mathcal G^\star$,
and make the children of $X_v^\star$ to be the original
children of $X_v$ in $\mathcal G$;
\item define the model $P(X_V, X_V^\star)$ to have the same factors
(\ref{eq:factorization}) as $P(X_V)$, but substituting every occurrence
of $X_v$ in $pa_{\mathcal G}(X_{D_i})$ by the corresponding
$pa_{\mathcal G^\star}(X_{D_i})$. Moreover, define $P_v^\star(X_v^\star\ |\ X_v)$
such that
\begin{equation}
\label{eq:artificial}
P_v^\star(X_v^\star = x\ |\ X_v = x) = 1
\end{equation}
\begin{equation}
\label{eq:final_fact}
P(X_V, X_V^\star) = \prod_{i = 1}^K P_i(X_{D_i}\ |\ pa_{\mathcal G^\star}(X_{D_i}) \backslash X_{D_i})
\prod_{X_v \in X_V}P_v^\star(X_v^\star\ |\ X_v)
\end{equation}
\end{itemize}

Since the last group of factors is identically equal to 1, they can be dropped from
the expression.

From (\ref{eq:artificial}), it follows that $P(X_V = x_V, X_V^\star =
x_V) = P(X_V = x_V)$. Since no two vertices in the same district can now have
a parent-child relation, all districts in $\mathcal G^\star$ are barren
and as such we can parameterize $P(X_V = x_V, X_V^\star = x_V)$
according to the results of the previous section. A similar trick was exploited by
\cite{silva:09} to reduce a problem of modeling ADMG probit
models to Gaussian models.

\begin{figure}
\begin{center}
\begin{tabular}{ccc}
\epsfig{file=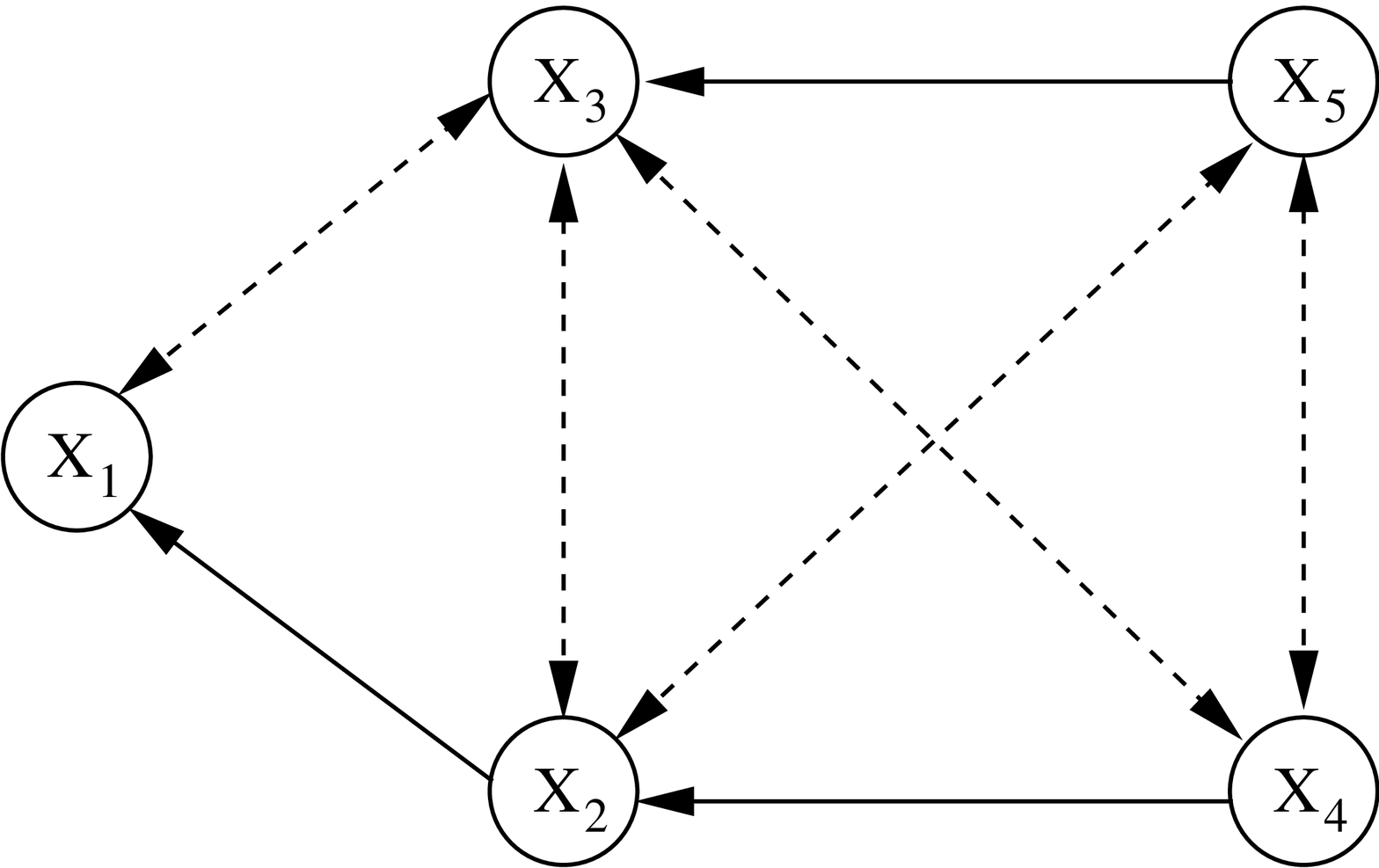,width=4.3cm} &
\epsfig{file=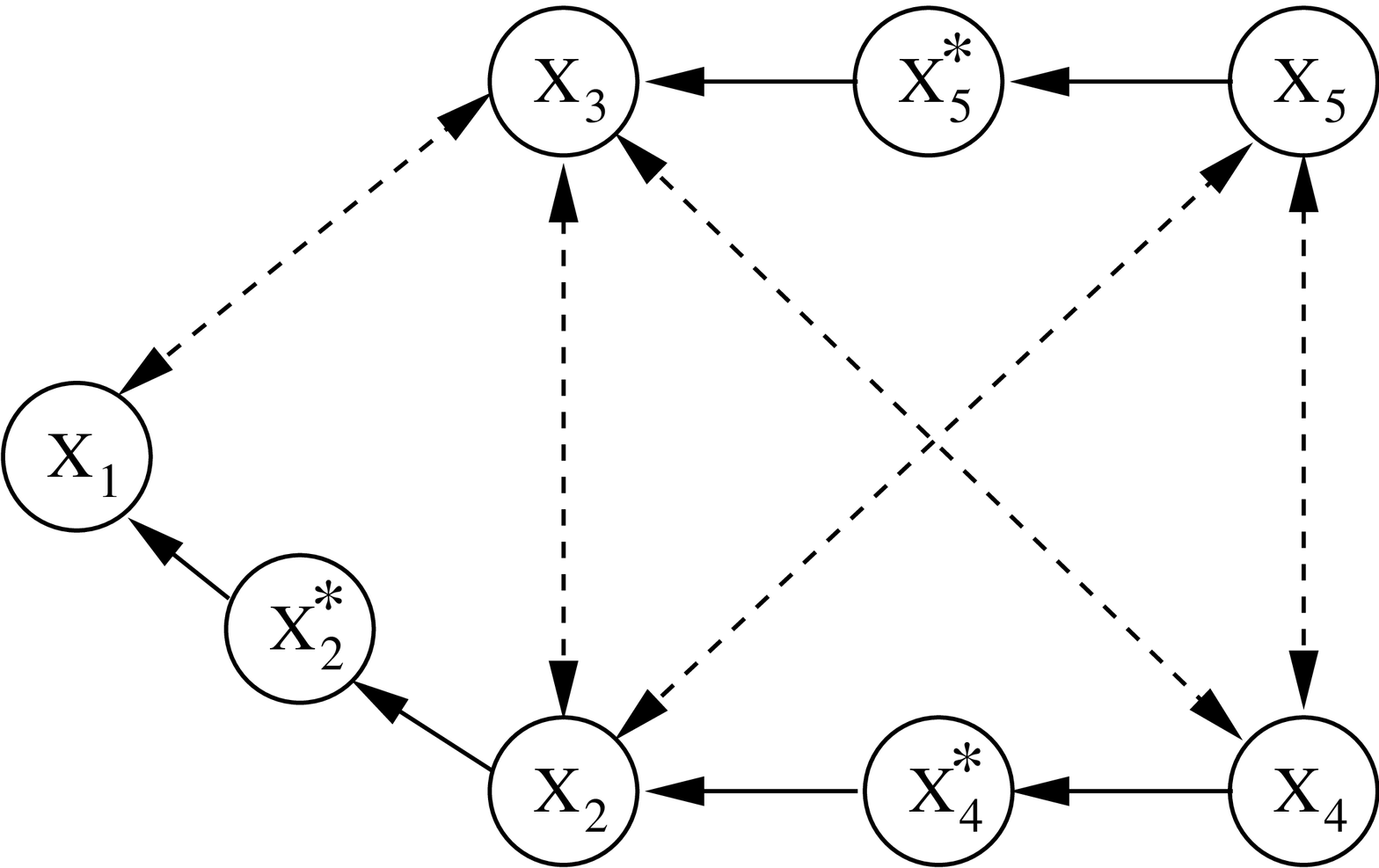,width=4.3cm} &
\epsfig{file=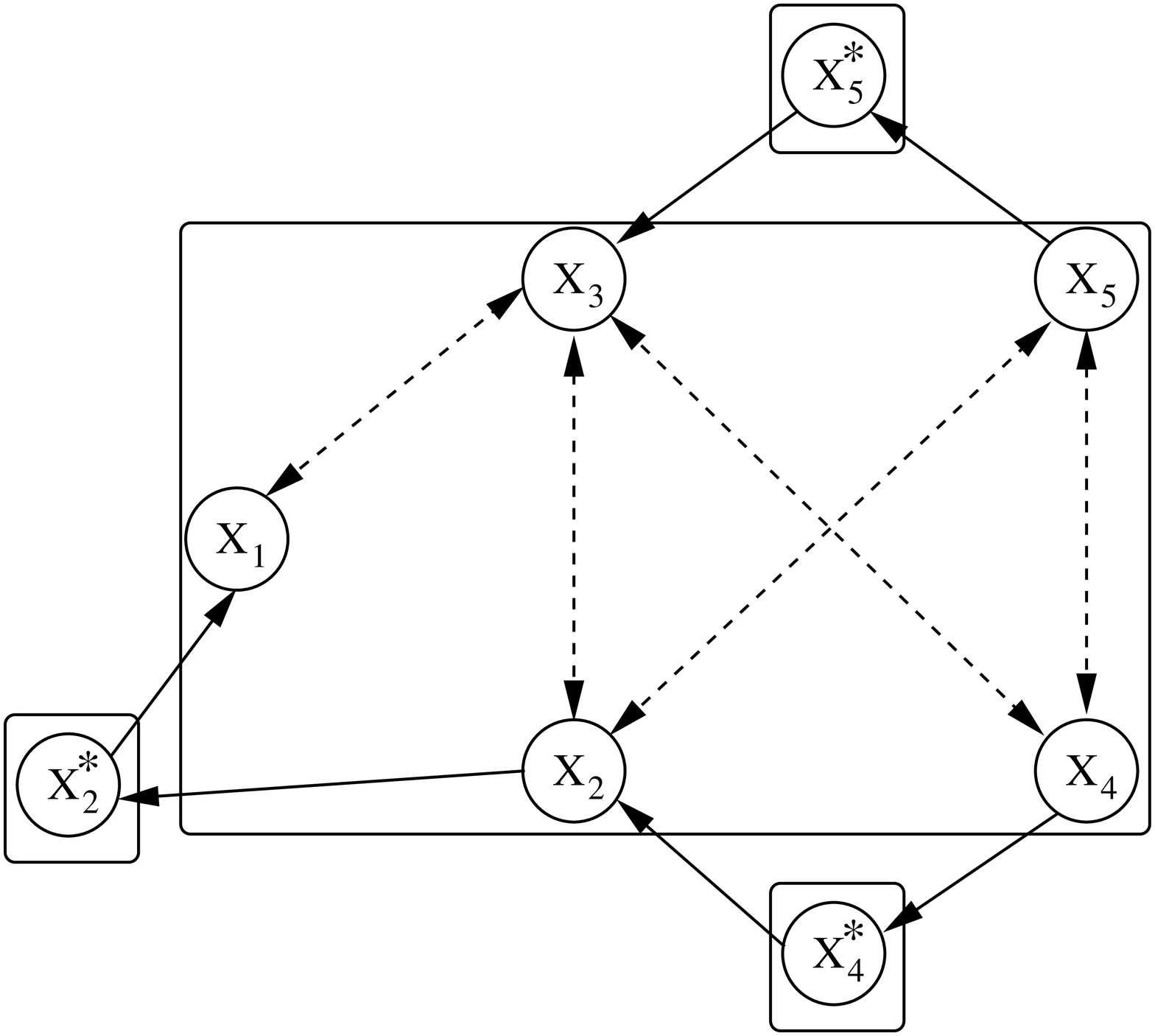,width=4.3cm} \\
(a) & (b) & (c)\\
\end{tabular}
\end{center}
\caption{(a) A mixed graph with a single district that includes all five vertices.
(b) The modified graph after including artificial vertices (artificial vertices for
childless variables are ignored). (c) A display of the four districts of the modified
graph in individual boxes. All districts are now barren, i.e., no directed edges can
be found within a district.}
\label{fig:reduction}
\end{figure}

Figure \ref{fig:reduction} provides an example, adapted from
\cite{richardson:09}. The graph has a single district containing all
vertices. The corresponding transformed graph generates several
singleton districts composed of one artificial variable either. In
Figure \ref{fig:reduction}(c), we rearrange such districts to
illustrate the decomposition described in Section \ref{sec:highlevel}.

\section{Copula MCDNs and Parameter Estimation}
\label{sec:copula}

The main result of Section \ref{sec:mcdn} is that we can parameterize
a MCDN model by parameterizing the factors in Equation
(\ref{eq:barren_fact}) corresponding to each district, which are then
tied together by the joint model (\ref{eq:final_fact}). However, we
have not yet specified how to construct each $F_S$ and $F_v$. In this
section, we describe a particularly convenient way of parameterizing
such factors. We introduce {\it copula MCDN models} -- a particular
instantiation of the MCDN family -- and how to estimate its
parameters.

Copulas are a flexible approach to defining dependence among a set
random variables. This is done by specifying the dependence structure
and the marginal distributions separately \cite{nelsen:07} (see also
\cite{kirshner:07} for a machine learning perspective). Simply put, a
copula function $C(u_1,\ldots,u_t)$ is just the CDF of a set of
dependent random variables, each with the uniform marginal
distribution over $[0,1]$. To define a joint distribution
over a set of variables $\{X_v\}$ with arbitrary marginal CDFs
$F_v(x_v)$, we simply transform each $X_v$ into a uniform variable
$u_v$ over $[0,1]$ using $u_v\equiv F_v(x_i)$.  The resulting joint
CDF $F(x_1,\ldots,x_t) = C(F_1(x_1),\ldots,F_t(x_t))$ incorporates
both the dependence encoded in $C$ and the marginal distributions
$F_v$.

Returning to ADMGs, let $\mathcal{G}_i$ be the subgraph corresponding
to a barren district $X_{D_i}$.  We parameterize a conditional CDF
$F_i(x_{D_i}|pa_{\mathcal{G}}(X_{D_i}))$ of form
\eqref{eq:barren_fact} Markov with respect to $\mathcal{G}_i$ by
defining the marginal CDFs and copula dependence separately.  In our
implementation the marginal probability for binary or ordinal $X_v$ is
an unconstrained conditional probability mass function.  The ordering
over the values of $X_v$, $\preceq$, naturally defines the marginal
$F_v(x_v|pa_\mathcal{G}(X_{v}))$:
\begin{equation}
\label{eq:cop_marg_discrete}
F_v(x_v\ |\ pa_{\mathcal G}(X_{v})) = \sum_{x \preceq x_v} \eta_{x}^{pa_{\mathcal G}(X_v)}
\end{equation}
\noindent where $\eta$ are the marginal parameters; conditioned upon
the parents of $X_v$, $\eta_x^{pa_{\mathcal G}(X_v)}$ is simply the
probability that $X_v = x$.  In our implementation for continuous
$X_v$, we define the marginal $F_v(x_v|pa_\mathcal{G}(X_{v}))$ using
conditional Gaussians:
\begin{equation}
\label{eq:cop_marg_continuous}
F_v(x_v\ |\ pa_{\mathcal G}(X_{v})) = \Phi(x_v; \ \textstyle \sum_{j=1}^K \eta_{vj}\phi_j(pa_\mathcal{G}(X_{v})), \sigma^2_v),
\end{equation}
with variance $\sigma_v^2$ and mean given by a linear regressor of
fixed basis functions $\phi_j(\cdot)$.

For a copula with the required bi-directed dependence among $X_{D_i}$,
we adopt the approach of product copulas \cite{liebscher:08}.  For
each clique $S$ in $\mathcal G_i$ let $C_S(u_S)$ be a
$|S|$-dimensional copula.  Let $d_v$ be the number of cliques variable
$X_v$ is in and define $a_v\equiv u_v^{1/(d_v+1)}$.  The product of
copulas given by:
\begin{align}
C_{D_i}(u_{D_i}) = \prod_{S\in{\mathcal{C}_i}}  C_S(a_S) \prod_{v\in D_i} a_v
\end{align}
can be shown to be a copula itself \cite{liebscher:08}.  Plugging in
the marginal distributions by defining $u_v\equiv F_v(x_v\ |\
pa_{\mathcal G}(X_{D_i}))$, the joint CDF over $x_{D_i}$ becomes:
\begin{equation}
\label{eq:cop_cop}
F_i(x_{D_i}\ |\ pa_{\mathcal G}(X_{D_i})) = 
           \left[\!\prod_{S \in \mathcal{C}_i} C_S(a_S)\!\right]
           \left[\!\prod_{v \in D_i}a_v\!\right] \quad\text{where $a_v\equiv F_v(x_v\ |\ {pa_{\mathcal G}(X_v)})^{1/(d_v+1)}$}
\end{equation}
The joint CDF has the form \eqref{eq:barren_fact} required to be
Markov with respect to $\mathcal G_i$.

We take an easy approach to parameter estimation commonly employed in
the copula literature:
\begin{enumerate}
\item fit the (conditional) marginals in (\ref{eq:cop_marg_discrete}) or (\ref{eq:cop_marg_continuous}) individually (by maximizing likelihood);
\item calculate the corresponding ``pseudodata'' $a_v$;
\item plug the estimated ``pseudodata'' into (\ref{eq:cop_cop}), and
      maximize the likelihood of the product copula \eqref{eq:cop_cop}. 
      Note that information from the parents has been absorbed into the calculation of $a_v$ via
      (\ref{eq:cop_marg_discrete}) or (\ref{eq:cop_marg_continuous}).
\end{enumerate}
Although the result is not a maximum likelihood estimator, it is a
practical procedure that does give consistent estimators
\cite{kirshner:07}. Given the pseudodata, the third 
step is maximum likelihood estimation of a CDN model as discussed by
\cite{huang:10}. In our implementation, used in Section
\ref{sec:experiments}, we substitute Step 3 by something even simpler to 
program\footnote{Maximum likelihood estimation requires the gradient
of the density with respect to the parameters. That is, we need
derivatives on top of the message-passing scheme that transforms a
CDF into a density function \cite{huang:10}.}, while providing a proof
of concept for the feasibility of Bayesian procedures: we put a prior
over the copula parameters and do Metropolis-Hastings (MH) with a
Gaussian random walk proposal. To calculate the MH ratio we only need
the likelihood function, which again can be obtained from the
message-passing scheme of \cite{huang:08,huang:10}.

\section{Experiments}
\label{sec:experiments}

We evaluate the usefulness of the MCDN formalism by comparing the
$K$-fold cross validated log-predictive probabilities of copula MCDNs
and DAGs on four data sets.  Two data sets are synthetic (from the
alarm \cite{binder:97} and insurance networks \cite{beinlich:89}) so
that the ground truth structure is known and we can compare against an
overparameterized DAG.  The non-synthetic data sets are both from the
UCI repository (the Wisconsin breast cancer and SPECTF data sets
\cite{wolberg:90, uci:2010}).  All data sets, except for the SPECTF
data set which is continuous, consist of ordinal or binary variables.

In our experiments, copula MCDNs are parameterized as described in
(\ref{eq:cop_marg_discrete}) or (\ref{eq:cop_marg_continuous}), and
(\ref{eq:cop_cop}).  We use Frank copulas, for computational
convenience, with Gaussian $\mathcal{N}(0,10)$ priors on their
parameters $\theta$.
\begin{table}
\center
\caption{\label{tab:results}Average difference in
log predictive per observations (in millibits) and standard errors.
\textbf{\# v} is the number of variables and \textbf{\# b} is the number of
bi-directed edges in the ADMG.
}
\begin{tabular}{r|c|c|l|r@{$\,\pm\,$}l}
\textbf{Data set} & \textbf{\# v}
                  & \textbf{\# b}
   & \textbf{Variables marginalized out}
   & $\mathbb{E}\left[\Delta_\text{DAG}\right]$
   & s.e.
    \\
\hline
Insurance & 25 & 2
    & {\scriptsize Driving Skill, Mileage}
    & 72.72 & 17.15
\\
Alarm & 33 & 4
    & {\scriptsize Err Cauter, TPR, KinkedTube, ArtCO2}
    & 76.27 & 13.78
\\
\hline
BreastCancer & 10 & 5
    & \emph{structure inferred}
    & 686.50 & 76.62
\\
SPECTF & 44 & 25
    & \emph{structure inferred}
    & -21.14 & 25.74
\end{tabular}
\end{table}
\begin{figure}
\begin{minipage}{.5\linewidth}
\centering
\includegraphics[width=\textwidth]{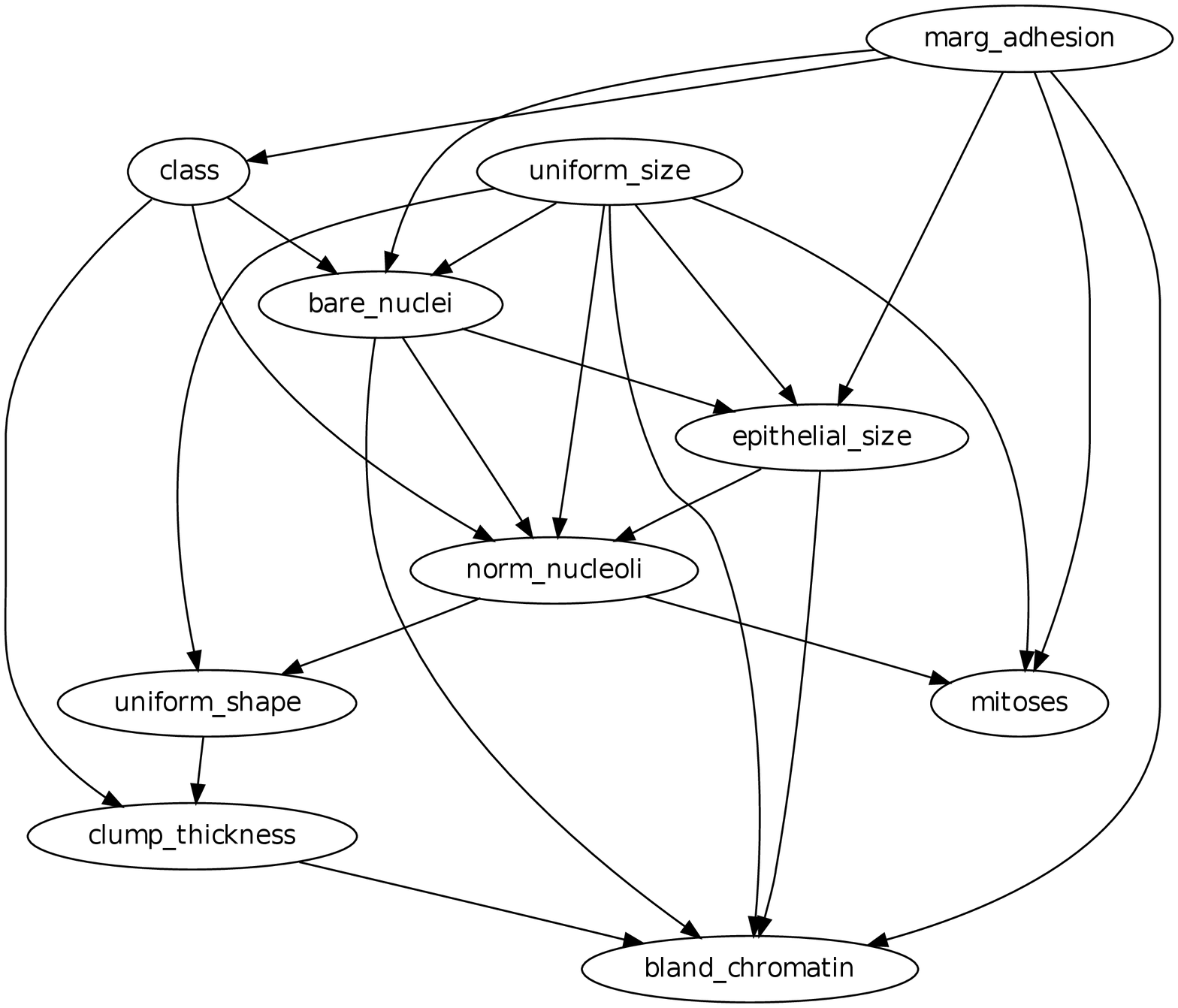}
\end{minipage}
\begin{minipage}{.5\linewidth}
\centering
\includegraphics[width=.95\textwidth]{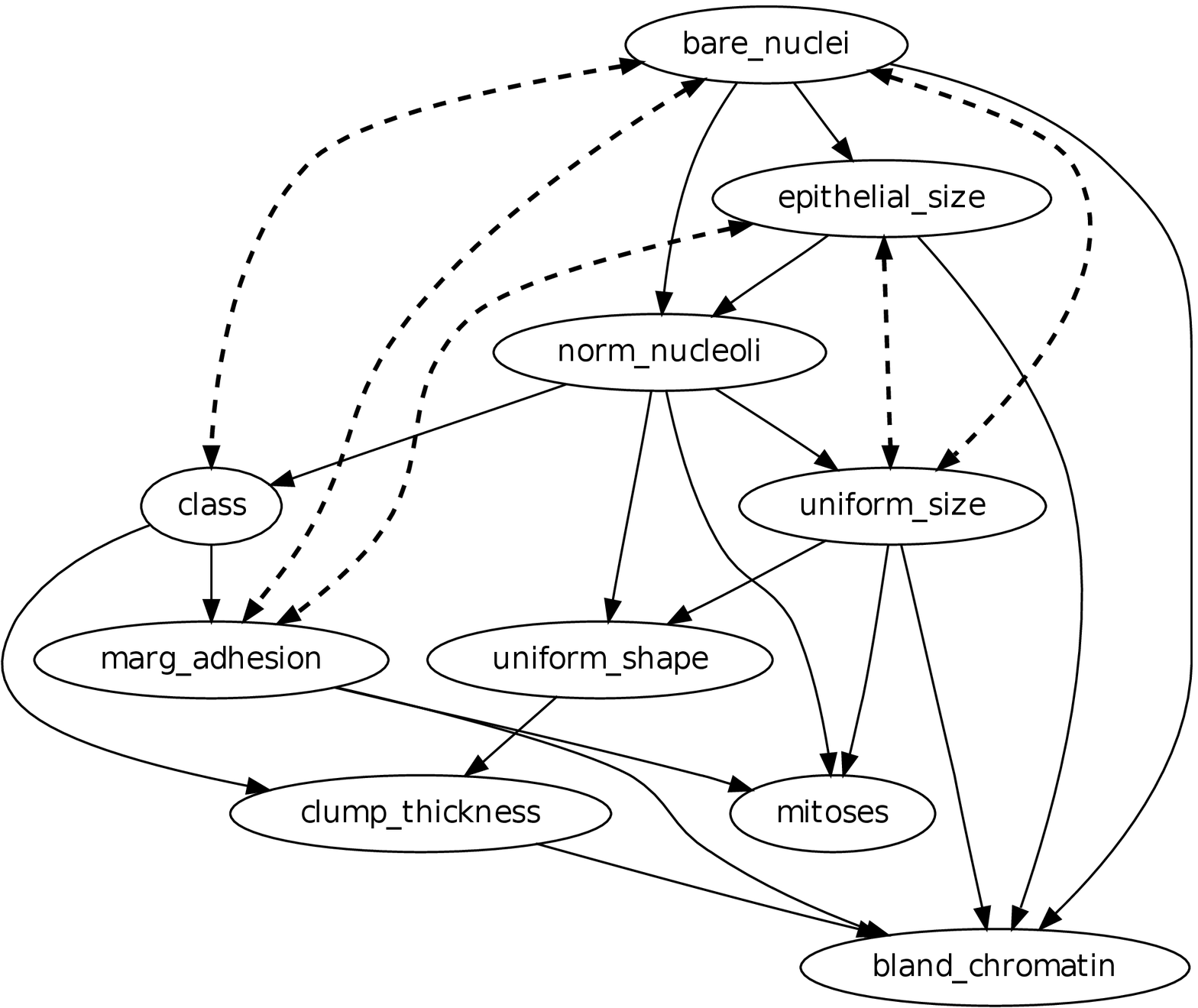}
\end{minipage}
\caption{\label{fig:cancerstruct}
DAG (left) and ADMG (right) structures inferred from the Wisconsin breast cancer data set.
}
\end{figure}
\paragraph{Known structure}
Several common cause variables (listed in table~\ref{tab:results})
were marginalized out of the data to introduce bi-directed edges to
the true structure.  An overparameterized DAG is able, parametrically,
to capture a broader set of conditional dependencies (by having
additional edges as well as broader parameterization) than those of a
copula MCDN; however it has many more parameters (exponential in the
parents of the district of the corresponding MCDN).  Hence we compare
these models on a small sample size of $300$.

The difference, in millibits, of the log predictive probability
between that of the copula MCDN and of the overparameterized DAG, per
cross-validation test set, is calculated as follows:
\begin{align*}
\Delta_{\text{DAG}}
    &= \frac{1000}{n_k} \left[\log_2 \tilde{p}(x_k|\mathcal{D}_k, \eta_k, \text{MCDN}) - \log_2 p(x_k|\mathcal{D}_k, \eta_k, \text{DAG})\right]
\end{align*}
\noindent where $x_k$ and $\mathcal{D}_k$ are the $k$th test and training set,
respectively, and $\eta_k$ are the maximum likelihood parameters of the
marginals from $\mathcal{D}_k$.

We calculate the predictive probability of the data set,
$\tilde{p}(x_k|\mathcal{D}_k,\eta_k,\text{MCDN})$, by averaging $p(x_k
| \mathcal{D}_k, \eta_k, \theta, \text{MCDN})$ over samples of the
copula parameter $\theta$. Positive $\Delta_\text{DAG}$ tells us on
average how many millibits better the prediction from the MCDN is over
the DAG model.  In both cases the log predictive probabilities were
significantly higher, although slight. Comparing to a DAG with
marginal parameters marginalized produced the same numbers (up to 5
s.f.) shown in table~\ref{tab:results}.

\paragraph{Unknown structure}

Next we ran an experiment on ordinal data without known structure.  We
used the original Wisconsin breast cancer data set from the UCI
repository \cite{wolberg:90}.  The ADMG and DAG structures shown in
figure~\ref{fig:cancerstruct} were inferred using MBCS*
\cite{pellet:08} and the $\chi^2$ test.  We then repeated the
procedure described above, instead calculating $\Delta_\text{DAG}$
relative to the inferred DAG rather than an overparameterized DAG, to
obtain the results also shown in table~\ref{tab:results}. On average,
the model performed encouragingly.

Finally, we used the SPECTF continuous data set from the UCI
repository \cite{uci:2010}.  We used this data in a more realistic
fashion: instead of learning the structure from the entire data set
then performing predictions of subsets, the structure learning is
incorporated into the $K$-fold cross validation.  We used $K=5$ for
this experiment and a score-based structure learning algorithm
\cite{schmidt:07} to find the DAG followed by fitting the bi-directed
edges using the residuals with the directed structure fixed.
Furthermore, if districts were not tree-structures, they were thinned
into trees (ordered by weakest residuals).  The residuals were fit by
testing marginal independence using \cite{gretton:07}.  This combined
technique allowed the structure to be inferred efficiently.

We compared this copula MCDN to a Gaussian DAG model (fit using just
the DAG learning algorithm of \cite{schmidt:07} and maximum
likelihood).  The results are shown in table~\ref{tab:results}.  The
number of bi-directed edges given is the average over the $K=5$ cross
validation folds.

In this case, the copula MCDN performed worse than the DAG model.
Note that the fitting procedure is suboptimal for MCDNs and, for
computational efficiency, does not alternate between learning directed
and bi-directed edges, and the bi-directed structure is limited to
tree-structured.  We also tried fitting a copula CDN, that is,
omitting the DAG search step and just fitting the residuals. Compared
to this model, the MCDN had an average difference of $11,504 \pm
2,456$ millibits suggesting that the DAG marginals are dominating the
copula MCDN fit on these data.

\section{Conclusion}
\label{sec:conclusions}

Acyclic directed mixed graphs are a natural generalization of DAGs.
While ADMGs date back at least to \cite{wright:21}, the potential of
this framework has only recently being translated into practical
applications due to advances into complete parameterizations of
Gaussian and discrete networks
\cite{richardson:02,drton:08,richardson:09}. The framework of
cumulative distribution networks \cite{huang:08,huang:10} introduced
new approaches for more constrained by widely applicable families of
marginal independence (bi-directed) models. By extending CDNs to the
full ADMG case, we expect that ADMGs will be readily accessible and as
widespread as DAG models.

There are several directions for future work. While classical
approaches for learning Markov equivalence classes of ADMGs have been
developed by means of multiple hypothesis tests of conditional
independencies \cite{sgs:00}, a model-based approach based on Bayesian
or penalized likelihood functions can deliver more robust learning
procedures and a more natural way of combining data with structural
prior knowledge. ADMG structures can also play a role in multivariate
supervised learning, that is, structured prediction problems. For
instance, \cite{silva:07c} introduced some simple models for
relational classification inspired by ADMG models and by the link to
seemingly unrelated regression \cite{zellner:62}. However, efficient
ADMG-structured prediction methods and new advanced structural
learning procedures will need to be developed.

\bibliographystyle{abbrv}
\small
\bibliography{../../../bib/rbas,cond_cdn}

\subsubsection*{APPENDIX A -- PROOFS}

{\bf Proposition 1.} {\it A probability function $P(X_V)$ is Markov
with respect to $\mathcal G$ if it can be factorized according to
(\ref{eq:factorization}), given that each $P_F(X_{D_i}\ |\
pa_{\mathcal G}(X_{D_i}) \backslash X_{D_i})$ is Markov with respect
to $\mathcal G_i$.}\\

Before we prove this theorem, we need to state the following result
from \cite{richardson:03}. Given an ancestral set $A$, the {\it Markov
  blanket} of vertex $X_v$ in $A$, $mb(X_v, A)$, is given by the
district of $X_v$ in $(\mathcal G)_A$ (except $X_v$ itself) along with
all parents of elements of this district.  Let a {\it total ordering}
$\prec$ of the vertices of $\mathcal G$ be any ordering such that if
$X_v \prec X_t$, then $X_t$ is not an ancestor of $X_v$ in $\mathcal
G$. A probability measure is said to satisfy the {\it ordered local
  Markov condition} for $\mathcal G$ with respect to $\prec$ if, for
any $X_v$ and ancestral set $A$ such that $X_t \in A\backslash\{X_v\} \Rightarrow 
X_t \prec X_v$, we have $X_v$ is independent of $A\backslash (mb(X_v, A) \cup \{X_v\})$
given $mb(X_v, A)$. The main result from \cite{richardson:03} states:\\

\noindent
{\bf Theorem 1.} {\it The ordered local Markov condition is equivalent to the global
Markov condition in ADMGs}\footnote{Notice this reduces to the standard
  notion of local independence in DAGs, where a vertex is independent
  of its (non-parental) non-descendants given its parents, from which
  d-separation statements can be derived
  \cite{lauritzen:96,pearl:88}.}.\\

\noindent {\it Proof of Proposition 1}: The proof is done by induction on $|X_V|$, with the case
$|X_V| = 1$ being trivial. We will show that if $P(X_V)$ is a probability function that
factorizes according to (\ref{eq:factorization}), as given by an ADMG
$\mathcal G$, then $P(X_V)$ is Markov with respect to $\mathcal G$.
To prove this, first notice there must be some $X_v$ with no children in
$\mathcal G$, since the graph is acyclic. Let $X_{D_i}$ be the
district of $X_v$. By assumption,
\begin{equation}
\label{eq:prepare_mb}
\begin{array}{rcl}
P(X_V) &=& P_F(X_v\ |\ X_{D_i} \cup pa_{\mathcal G}(X_{D_i})) \times
P_F(X_{D_i}\backslash X_v\ |\ pa_{\mathcal G}(X_{D_i})\backslash X_{D_i})\\
&\times& \prod_{j \ne i} P_F(X_{D_j}\ |\ pa_{\mathcal G}(X_{D_j})\backslash X_{D_j})
\end{array}
\end{equation}
Since $X_v$ is childless, it does not appear in any of the factors in the expression 
above, except for the first. Hence, 
\begin{equation}
\label{eq:recurse_fact}
P(X_V \backslash X_v) = 
P_F(X_{D_i}\backslash X_v\ |\ pa_{\mathcal G}(X_{D_i})\backslash X_{D_i})
\times \prod_{j \ne i} P_F(X_{D_j}\ |\ pa_{\mathcal G}(X_{D_j})\backslash X_{D_j})
\end{equation}
which by induction hypothesis is Markov with respect to the marginal
graph $(\mathcal G)_{X_V \backslash X_v}$ (one minor detail is that
$(\mathcal G)_{X_V \backslash X_v}$ might have more districts than
$\mathcal G$ after removing $X_v$. However, the result still holds by
further factorizing $P_F(X_{D_i}\backslash X_v\ |\ pa_{\mathcal
G}(X_{D_i})\backslash X_{D_i})$ according to the newly formed
districts of $X_{D_i}\backslash X_v$ -- which is possible by the construction of
$P_F(\cdot)$ and $\mathcal G_i$). By the ordered local Markov property
for ADMGs and any ordering $\prec$ where $X_v$ is the last vertex,
probability function $P(X_V)$ will be Markov with respect to $\mathcal
G$ if, according to $P(X_v)$, the Markov blanket of $X_v$ in $\mathcal
G$ makes $X_v$ independent of the remaining vertices. But this true by
construction, since this Markov blanket is contained in $X_{D_i} \cup
pa_{\mathcal G}(X_{D_i})$ according to Theorem 1. $\Box$\\

Notice that factorization (\ref{eq:factorization}) is seemingly
cyclical: for instance, Figure \ref{fig:district}(a) implies the
factorization $P_F(X_1, X_2\ |\ X_4)P_F(X_3, X_4\ |\ X_1)$. This
suggests that there are additional constraints tying parameters
across different factors. However, there are no such constraints,
as guaranteed through the following result:\\
                                                          
\noindent {\bf Proposition 2.} {\it Given an ADMG $\mathcal G$ with
  respective subgraphs $\{\mathcal G_i\}$ and districts $\{X_{D_i}\}$,
  any collection of probability functions $P_F(X_{D_i}\ |\
  pa_{\mathcal G}(X_{D_i}) \backslash X_{D_i})$, Markov with respect
  to the respective $\mathcal G_i$, implies that
  (\ref{eq:factorization})
  is a valid probability function (a non-negative function that integrates to 1).}\\

\noindent {\it Proof}: It is clear that (\ref{eq:factorization}) is
non-negative.  We have to show it integrates to 1. As in the proof of
Proposition 1, first notice there must be some $X_v$ with no children
in $\mathcal G$, since the graph is acyclic. Those childless vertices
can be marginalized as in Equation (\ref{eq:recurse_fact}) if they do
not appear on the right-hand side of any factor $P_F(\cdot\ |\
\cdot)$, and removed from the graph along with all edges adjacent to
them. After some marginalizations, suppose that in the current
marginalized graph, a childless vertex $X_\emptyset$ appears on the
right-hand side of some factor $P_F(X_{D_i}\ |\ pa_{\mathcal
  G}(X_{D_i}) \backslash X_{D_i})$. Because $X_\emptyset$ has no
children in $X_{D_i}$, by construction $X_{D_i}$ are $X_v$ are
independent given the remaining elements in $pa_{\mathcal G}(X_{D_i})
\backslash X_{D_i}$.  As such, $X_\emptyset$ can be removed from the
right-hand side of all remaining factors, and then marginalized. The
process is repeated until the last remaining vertex is marginalized,
giving 1 as the result. $\Box$.\\

\noindent {\bf Proposition 3.} {\it $F_{i}(x_{D_i}\ |\ pa_{\mathcal
    G}(X_{D_i}))$ is a CDF for any choice of $\{\{F_S(x_S\ |\
  pa_{\mathcal G}(X_S)\}, \{F_v(x_v\ |\ pa_{\mathcal G}(X_v))\}\}$.
  If, according to each $F_S(x_S\ |\ \cdot)$, $X_s \in X_S$ is
  independent of any element in $pa_{\mathcal G}(X_{D_i})\backslash
  pa_{\mathcal G}(X_s)$, the corresponding conditional probability
  function $F_{i}(x_{D_i}\ |\ pa_{\mathcal G}(X_{D_i}))$ is Markov
  with respect to
  $\mathcal G_i$.}\\

\noindent {\it Proof:} Each factor in (\ref{eq:barren_fact}) is a CDF with
respect to $X_{D_i}$, with $pa_{\mathcal G}(X_{D_i})$ fixed, and hence
its product is also a CDF \cite{huang:08}. To show the Markov
property, it is enough to consider the modified graph $\mathcal G_i'$
constructed by transforming all directed edges in $\mathcal G_i$ into
bi-directed edges, since the implied distributions conditional on
$pa_{\mathcal G}(X_{D_i})$ for $\mathcal G_i'$ and $\mathcal G_i$ are
Markov equivalent \cite{richardson:02}.  It follows directly from the
assumptions and the properties of CDFs that disconnected sets in
$\mathcal G_i'$ are marginally independent, which corresponds to the
Markov properties of bi-directed graph $\mathcal G_i'$
\cite{richardson:03}. $\Box$\\

\subsubsection*{APPENDIX B -- BINARY CASE: RELATION TO COMPLETE PARAMETERIZATION}

A complete parameterization for binary ADMG models is described by
\cite{richardson:09}. As we will see, 
parameters are defined in the context of different marginals,
analogous to the purely bi-directed case \cite{drton:08}.

As in the bi-directed case, the joint probability distribution is given
by an inclusion-exclusion scheme:
\begin{equation}
\label{eq:admg_complete}
\displaystyle
P(X_V = \alpha(V)) = \sum_{C:\alpha^{-1}(0)\subseteq C \subseteq V}
(-1)^{|C\backslash \alpha^{-1}(0)|}\prod_{H \in [C]_{\mathcal G}}
P(X_H = 0\ |\ X_{tail(H)} = \alpha(tail(H)))
\end{equation}
\noindent where $\alpha(V)$ is a binary vector in $\{0, 1\}^{|X_V|}$
and $\alpha^{-1}(0)$ is a function that indicates which elements
in $X_V$ were assigned to be zero.

Each $C$ indicates which elements are set to zero in the respective
term of the summation. Depending on $C$, the factorization
changes. $[C]_{\mathcal G}$ is a set of subsets of $X_V$: one subset
per district, each subset being barren in $\mathcal G$. The
corresponding $tail(H)$ is the Markov blanket for the ancestral set
that contains $H$ as its set of childless vertices.

As in our discussion of standard CDNs, Equation
(\ref{eq:admg_complete}) can be interpreted as the CDF-to-probability
transformation (\ref{eq:cdf2pmf}). It can be rewritten as
\[
\begin{array}{rcl}
\displaystyle
P(X_V = \alpha(V)) & =& \sum_{C:\alpha^{-1}(0)\subseteq C \subseteq V}
(-1)^{|C\backslash \alpha^{-1}(0)|} \times \\
&& \displaystyle
\prod_{H \in D_i \cap [C]_{\mathcal G}}
P(X_{D_i}\backslash tail(H) \le \alpha(V)\ |\ X_{tail(H)} = \alpha(tail(H)))
\end{array}
\]
Hence, this parameterization can also be interpreted as a CDF
parameterization.  One important difference is that each term in the
summation uses only a subset of each district, $X_{D_i}\backslash
tail(H)$ instead of $X_{D_i}$. Notice that some elements of $X_{D_i}$
appear in the conditioning set (i.e., $tail(H)$ contains some of the remaining 
elements of $X_{D_i}$, on top of the respective parents).

The need for using subsets comes from the necessity of enforcing
independence constraints entailed by bi-directed paths. As in the CDN
model, the MCDN criterion factorizes each CDF according to its cliques as
an indirect way of accounting for such constraints. Hence, we do not
construct factorizations for different marginals: each factor within a
summation term in (\ref{eq:admg_complete}) includes all elements of
each district. We enforce that they remain barren by the
transformation in Section \ref{sec:general} $-$ which is unnecessary
in \cite{richardson:09} because only barren subsets are being
considered.

To understand how the parameterizations coincide, or which constraints
analogous to (\ref{eq:min_constraint}) emerge in our parameterization,
consider first the following example. Using the results from
\cite{richardson:09}, the graph in Figure
\ref{fig:district}(a) needs the specification of the following
marginals:
\begin{equation}
\label{eq:ex_factorize}
\begin{array}{rcl}
P(X_1, X_4) & = & P(X_1)P(X_4)\\
P(X_1, X_3, X_4) & = & P(X_3, X_4\ |\ X_1)P(X_1)\\
P(X_1, X_2, X_4) & = & P(X_1, X_2\ |\ X_4)P(X_4)\\
P(X_1, X_2, X_3, X_4) & = & P(X_1, X_2\ |\ X_4)P(X_3, X_4\ |\ X_1)\\
P(X_1, X_3) & = & P(X_3\ |\ X_1)P(X_1)\\
P(X_2, X_4) & = & P(X_2\ |\ X_4)P(X_4)\\
\end{array}
\end{equation}
As an example, the probability $P(X_{14} = 0, X_{23} = 1) \equiv P(X_1 = 0, X_2 = 1, X_3 = 1, X_4 = 0)$
can be derived from the above factorizations and (\ref{eq:admg_complete}) as

$P(X_1 = 0, X_2 = 1, X_3 = 1, X_4 = 0)$
\[
\begin{array}{rcl}
& = & P(X_1 \le 0, X_2 \le 1, X_3 \le 1, X_4 \le 0) - P(X_1 \le 0, X_2 \le 1, X_3 \le 0, X_4 \le 0) -\\
&   & P(X_1 \le 0, X_2 \le 0, X_3 \le 1, X_4 \le 0) + P(X_1 \le 0, X_2 \le 0, X_3 \le 0, X_4 \le 0)\\
&&\\
& = & P(X_1 = 0, X_4 = 0) - P(X_1 = 0, X_3 = 0, X_4 = 0) -\\
&   & P(X_1 = 0, X_2 = 0, X_4 = 0) + P(X_1 = 0, X_2 = 0, X_3 = 0, X_4 = 0)\\
&&\\
& = & P(X_1 = 0)P(X_4 = 0) - P(X_{34} = 0\ |\ X_1 = 0)P(X_1 = 0) -\\
&   & P(X_{12} = 0\ |\ X_4 = 0)P(X_4 = 0) + P(X_{12} = 0\ |\ X_4 = 0)P(X_{34} = 0\ |\ X_1 = 0)
\end{array}
\]
\noindent where the last line comes from the pool of possible factorizations (\ref{eq:ex_factorize}). 
The corresponding probability using the MCDN parameterization is
\[
\begin{array}{rcl}
& = & P(X_1 = 0, X_2 = 1\ |\ X_4 = 0)P(X_3 = 1, X_4 = 0\ |\ X_1 = 0)\\
&&\\
& = & (P(X_1 \le 0, X_2 \le 1\ |\ X_4 = 0) - P(X_1 \le 0, X_2 \le 0\ |\ X_4 = 0)) \times\\
&&    (P(X_3 \le 1, X_4 \le 0\ |\ X_1 = 0) - P(X_3 \le 0, X_4 \le 0\ |\ X_1 = 0))\\
&&\\
& = & (P(X_1 = 0 |\ X_4 = 0) - P(X_1 = 0, X_2 = 0\ |\ X_4 = 0)) \times\\
&&    (P(X_4 = 0\ |\ X_1 = 0) - P(X_3 = 0, X_4 = 0\ |\ X_1 = 0))\\
&&\\
& = & (P(X_1 = 0) - P(X_1 = 0, X_2 = 0\ |\ X_4 = 0)) \times\\
&&    (P(X_4 = 0) - P(X_3 = 0, X_4 = 0\ |\ X_1 = 0))\\
&&\\
& = & P(X_1 = 0)P(X_4 = 0) - P(X_{34} = 0\ |\ X_1 = 0)P(X_1 = 0) -\\
&   & P(X_{12} = 0\ |\ X_4 = 0)P(X_4 = 0) + P(X_{12} = 0\ |\ X_4 = 0)P(X_{34} = 0\ |\ X_1 = 0)
\end{array}
\]
\noindent where the first line comes from the factorization of
$P(X_1 = 0, X_2 = 1, X_3 = 1, X_4 = 0)$ according to
(\ref{eq:factorization}) and the fourth line comes from the Markov
properties of each $\mathcal G_i$ factor. Although these
parameterizations have the same high-level parameters, they still do
not coincide, as shown in the next example.

For a more complicated case where an extra constraint appears in our parameterization,
consider Figure \ref{fig:reduction}(a). In \cite{richardson:09}, it is
shown that one of the parameters of the complete parameterization is
$P(X_1 = 0, X_3 = 0\ |\ X_2 = 0, X_4 = 0, X_5 = 0)$, which reflects
the fact that $X_1$ and $X_5$ are dependent given all other
variables. This also true in our case, except that according to Figure
\ref{fig:reduction}(c), our corresponding CDF is given by
\begin{equation*}
F(x_1\ |\ X_2)F(x_1, x_3)F(x_2, x_3)F(x_3, x_4)F(x_4, x_5)
F(x_3\ |\ X_5)F(x_2\ |\ X_4)
\end{equation*}
\noindent which, evaluated at $X_{12345} = 0$, gives
\[
P(X_1 = 0\ |\ X_2 = 0)P(X_1 = 0, X_3 = 0)P(X_2 = 0, X_3 = 0)P(X_3 = 0, X_4 = 0)\times
\]
\[
P(X_4 = 0, X_5 = 0)P(X_3 = 0\ |\ X_5 = 0)P(X_2 = 0\ |\  X_4 = 0)
\]
\noindent implying that $P(X_{12345} = 0)$ factorizes as
$f(X_1, X_2, X_3, X_4)g(X_2, X_3, X_4, X_5)$, the generalization to
(\ref{eq:min_constraint}).

\end{document}